\pdfoutput=1

\documentclass[11pt]{article}

\usepackage{acl}

\usepackage{times}
\usepackage{latexsym}
\usepackage[colorinlistoftodos,prependcaption,textsize=tiny]{todonotes}

\usepackage[T1]{fontenc}

\usepackage[utf8]{inputenc}

\usepackage{microtype}
\usepackage{multirow}

\usepackage{booktabs}

\usepackage{makecell}
\usepackage{subcaption}

\usepackage{etoolbox,siunitx}
\renewcommand{\bfseries}{\fontseries{b}\selectfont}
\newrobustcmd{\B}{\bfseries}

\usepackage{enumitem}
\usepackage{tablefootnote}

\newcommand{\Hquad}{\hspace{0.5em}} 

\def\algo{\textsc{TabT5}}
\def\wikisql{\textsc{WikiSQL}}
\def\totto{\textsc{ToTTo}}

\def\finqa{\textsc{FinQA}}

\def\scoder{\textsc{Enron}}
\def\gold{\textsc{Gold}}

\title{Table-To-Text generation and pre-training with \algo{}}

\author{Ewa Andrejczuk\thanks{\ \ corresponding authors}, Julian Martin Eisenschlos$^*$\thanks{\ \ main advisor}, \\ \textbf{Francesco Piccinno}, \textbf{Syrine Krichene}, \textbf{Yasemin Altun} \\ \\
Google Research, Z\"urich\\
\texttt{\{ewaa,eisenjulian,piccinno,syrinekrichene,altun\}@google.com}
}

\begin{document}
\maketitle
\begin{abstract}
Encoder-only transformer models have been successfully applied to different table understanding tasks, as in TAPAS~\cite{tapas}.
A major limitation of these architectures is that they are constrained to classification-like tasks such as cell selection or entailment detection. 
We present \algo{}, an encoder-decoder model that generates natural language text based on tables and textual inputs. 
\algo{} overcomes the encoder-only limitation by incorporating a decoder component and leverages the input structure with table specific embeddings and pre-training. 
\algo{} achieves new state-of-the-art results on several domains, including spreadsheet formula prediction with a 15\% increase in sequence accuracy, QA with a 2.5\% increase in sequence accuracy and data-to-text generation with a 2.5\% increase in BLEU.
\end{abstract}

\section{Introduction}


\begin{figure*}[ht]
    \centering
    \includegraphics[width=1.0\linewidth]{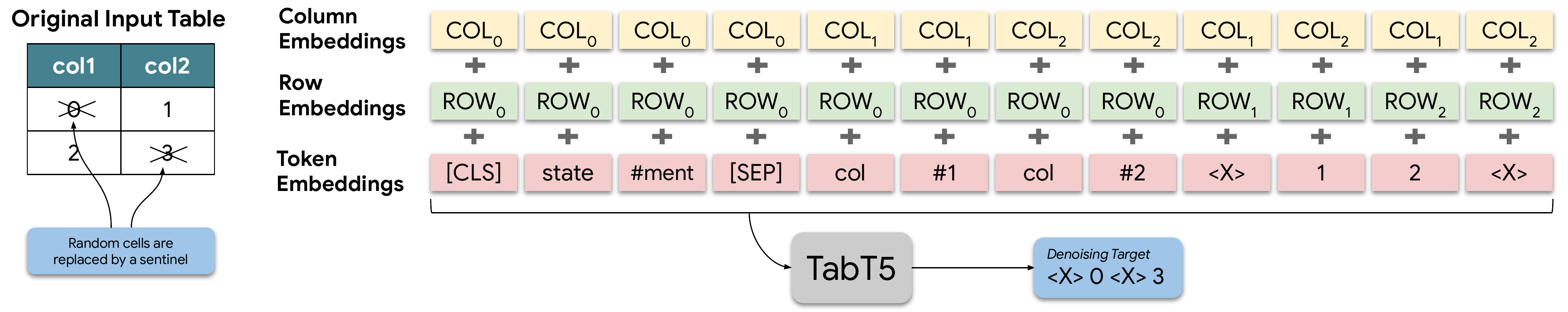}
    \caption{\algo{} linearizes the input table row by row and adds column and row embeddings to encode the 2-dimensional coordinates of each cell. The model is pretrained by randomly replacing 15\% of cells by a \texttt{<X>} marker and training the decoder to predict the hidden output in sequence.}
    \label{fig:model}
\vspace*{-5mm}
\end{figure*}

Large language models (LLMs) such as BERT~\cite{bert} or T5~\cite{t5} have shown impressive abilities to encode and generate fluent and coherent natural language text~\cite{lan2019albert,gururangan2020don,conneau2019unsupervised}. 
%
However their representation and generational capabilities are limited when it comes to structured or semi-structured domains like tables. 
This is mainly due to two reasons: (i) LLMs are only pre-trained on large amount of unstructured data (e.g., documents, news, etc.);  (ii) their underlying model architecture lacks a way to fully leverage this structure information.

Yet, structured and semi-structured data is ubiquitous on the web (e.g. web tables, database tables, PDF tables, spreadsheets store rich numerical information and provide concise summaries of data), and widely studied 
%
in the academia \cite{chen2021spreadsheetcoder,cheng2021fortap,totto,zhiruo2021tuta} and the industry (e.g. formula prediction in
Excel\footnote{\url{https://www.microsoft.com/excel}} and Google Sheets\footnote{\url{https://www.google.com/sheets/about}}, or extracting data from tables in Text-to-Speech Assistants).

Recently, several solutions propose to alleviate aforementioned issue by introducing pre-training or intermediate training strategies for tables.
For instance, \citet{tapas} propose to use a Masked Language Model (MLM) as pre-training objective to improve the contextual representation of BERT \cite{bert} over table inputs. 
To train their model, they introduce additional input embeddings that help the model understand the table structure. 
%
%
These pre-training models are designed and evaluated on datasets where the answers contain only table cells or aggregations of multiple cells, and not full sentences. 
In this paper, we tackle a set of distinct, complex tasks such as question answering and formula prediction that require full generation capabilities.

\noindent
In particular, our contributions are as follows:
\begin{itemize}[noitemsep,topsep=0pt,leftmargin=*]
    \item We present an encoder-decoder based model \algo{} (Table-and-Text-to-Text Transfer Transformer) that can be applied to data-to-text generation tasks by relying on special embeddings of the input structure. 
    \item We introduce different pre-training strategies that leverage web data containing tables.
    \item We evaluate our approach on four different table and text datasets in English and obtain state-of-the-art performance on several domains.
\end{itemize}


\section{Problem Definition}

The objective of our model is to learn a conditional sequence generator $P(y|x)$ where $x$ is endowed with extra two-dimensional structure. 
To encode said structure, each instance of $x$ is as a variable length sequence of tuples ${\left(u_i, t_i, c_i, r_i\right)}_{i=1}^N$ representing the \emph{components} of $x$.
In each component, $u_i$ is a natural language utterance, $t_i$ is the discrete type of the component (i.e. could be \emph{Question, Document Title, Table Caption, Table Header, Table Cell}, etc.), and $c_i$ and $r_i$ represent the two dimensional column and row coordinates for this component. 
This approach is general to represent the information layout in web documents and in particular tables where each table cell and each piece of metadata can map into a single component.
%

\section{Related Work}

\noindent
\textbf{Table language models.} 
Several works 
%
use a common serialization where table contents are linearized row by row~\cite{totto,zhiruo2021tuta,iida2021tabbie,mate,tapas}.
Another design choice is to use structural positional encoding, in addition to $1$-D encoding, to represent two
dimensional information such as the row and column positions of  tokens~\cite{tapas,mate,iida2021tabbie,zhiruo2021tuta}. 
An alternative 
is the the use of a structure-aware attention, in
contrast to a standard self-attention mechanism, to better leverage
the table structure~\cite{mueller-etal-2019-answering,tableformer}.
All of these models are encoder-only. 
Concurrent with our work ~\citet{gentap} propose a similar method to adapt to T5 to tabular data, however their pretraining approach relies on existing annotated datasets and focuses solely on QA applications.

\noindent
\textbf{Table Pre-training.}
Most pretraining methods 
follow the Masked Language Modeling (MLM) scheme, where some percentage of input tokens are randomly masked and successively predicted in an encoder only setup \cite{tapas,mate,tableformer}.
Some approaches \cite{zhiruo2021tuta,yin2020tabert,iida2021tabbie} apply the masking on a cell-level, where the full contents of a given cell is masked and then predicted. 
Our work differs in training the encoder and decoder jointly by using a de-noising scheme similar to the one used in T5~\cite{t5}. 

\noindent
\textbf{Table QA.} Given an input table, the task consists in producing an answer to a natural language question. We focus on \wikisql{}~\cite{wikisql}, and learn an encoder-decoder model with row/column embeddings in the weakly supervised setting without logical forms. \citet{tapas} use a similar approach with a BERT encoder-only model, while \citet{liu2022tapex} use a BART encoder-decoder model without extra embeddings.

\noindent
\textbf{Formula prediction.}
The task is to predict formula conditioned on headers and other contextual information, without an explicit natural language question.
\citet{chen2021spreadsheetcoder} propose to use a BERT-based architecture to compute an input header and a cell data vector that are fed to a two-step LSTM decoder. The decoder proposes a formula sketch and refines it with cell ranges.
\citet{cheng2021fortap} propose a similar approach where the representation of the target cell output by a table encoder~\cite{zhiruo2021tuta} is an input to a two-step LSTM-based decoder.
Our approach is simpler as a single model is used to solve the task end to end.

\noindent
\textbf{Data-to-Text.} The task consists in generating a natural language description given structured data input. 
\citet{totto} employ an encoder-decoder model where the encoder and decoder are both initialized with BERT~\cite{bert}. \citet{kale2020text} use a T5 model. In both, tables are linearized with row/column separator tokens. Our work differs as we use row/column embeddings, and we employ two pretraining schemes.

\section{\algo{} Model}

\algo\ uses the T5 pre-trained language model as a baseline architecture.
We linearize the table into a sequence of words, split words into word pieces (tokens) and concatenate the question and table tokens to create the input sequence. We include in the model row and column embeddings to encode table structure \cite{tapas}.  We add them on top of the token embeddings as model inputs and optimize them during training (Figure~\ref{fig:model}). 
The target sequence is a free-form answer. This can be an answer to a question for question-answering tasks, a table summary when no question is specified or a formula for the formula prediction tasks.

\section{Pre-training}


As a starting point, we use publicly available T5 checkpoints released by \citet{t5}. Next, we pre-train \algo\ on Wikipedia tables. We use the pre-training dataset proposed by \citet{tapas} which contains 6.2M tables (3.3M of class Infobox\footnote{\href{https://en.wikipedia.org/wiki/Help:Infobox}{https://en.wikipedia.org/wiki/Help:Infobox}} and 2.9M of class WikiTable). We also extract related passages that caption the table. We define two pre-training strategies described below.

\subsection{Denoising} \label{sec:MLM}

We design a denoising strategy for table-like data, following the method used in T5~\cite{t5}, by training the model to predict a target sequence containing the missing or corrupted tokens in the input table.
The target consists of all of the dropped-out spans of tokens, delimited by the sentinel token used in the input sequence (Figure~\ref{fig:model}). 
We replace 15\% of table cells and columns in the input with a mask token\footnote{\citeauthor{t5} experimentally show that 15\% corruption rate works best. We use the same rate for our denoising objective.}. 
This helps the model capture relationships between the neighbouring cells and between the related text. 
%

\subsection{ToTTification}

We define another pre-training strategy using the same Wikipedia tables (Section~\ref{sec:MLM}) inspired by ToTTo~\cite{totto}, to be used after denoising. For each table, we retrieve the statements that are in the same page as the table or link to the table page. We only keep statements that have an entity (Wikipedia URL, number or date) that matches the table, 4M in total. These statements become our target text. We add the matching entities in those statements as a (comma separated) plain text component of the input to guide the generation.

\begin{table*}[!t]
\centering
\footnotesize
\begin{tabular}{@{}lS[table-format=2.2(1), separate-uncertainty, detect-weight]S[table-format=2.2(1), separate-uncertainty, detect-weight]S[table-format=2.2(1), separate-uncertainty, detect-weight]S[table-format=2.2(1), separate-uncertainty, detect-weight]@{}} 
  & \multicolumn{2}{c}{\emph{Overall}} & \multicolumn{2}{c}{\emph{Non-Overlap}} \\

\toprule
 \textbf{Model -- Dev Set} & \textbf{BLEU} & \textbf{PARENT} & \textbf{BLEU} & \textbf{PARENT} \\
\midrule
\citeauthor{kale2020text} & 47.70 & 57.10 & 39.60 & 52.60 \\
T5-base & 47.00(43) & 55.96(31) & 38.50(48) & 51.14(33) \\
\hline
\algo-small & 47.80(26) & 56.89(29) & 39.30(26) & 51.93(35)\\
\algo-base  & 49.00(07) & 57.70(11) & 40.90(13) & 53.12(18)\\
\Hquad \textsc{+ToTTify} & \B 49.50(07) & \B 57.95(05) & \B 41.60(05) & \B 53.65(07)\\
\Hquad \textsc{-denoising} & 47.50(43) & 56.11(40) & 39.00(64) & 51.06(51) \\
\Hquad \textsc{-embeddings} & 48.60(17) & 57.12(23) & 40.50(26) & 52.71(28) \\
\algo-large  & 48.50(13) & 56.98(25) & 41.05(10) & 52.95(24)\\
\toprule
 \textbf{Model -- Test Set} & \textbf{BLEU} & \textbf{PARENT} & \textbf{BLEU} & \textbf{PARENT} \\
\midrule
\citeauthor{totto} & 44.00 & 52.60 & 35.10 & 46.80 \\
T5-base & 47.10 & 56.17 & 38.70 & 51.39 \\
\hline
\algo-base  & 48.80 & 57.60 &  40.70 & 53.20\\
\Hquad \textsc{+ToTTify} & \B 49.20 & \B 57.25 & \B 41.00 & \B 52.78\\
\bottomrule
\end{tabular}
	\caption{Text generation results for \totto{} on development (dev) and test sets. The \emph{Non-Overlap} set features examples that are out-of-domain from the training set. \algo{} provides improvements over existing approaches and \textsc{ToTTify} pretraining provides additional gains.}
	\label{table:tottoresults}
	\vspace*{-5mm}
\end{table*}

\section{Experiments}\label{sec:experiments}

In this section, we discuss the experiments we performed to show the effectiveness of our method. 

\subsection{Datasets}


\noindent \textbf{\wikisql} \cite{wikisql} is a Table-QA dataset containing $80.654$ instances. To create the dataset, crowd workers paraphrase a template-based question into natural language. Two other crowd workers' groups then verify and correct the quality of the proposed paraphrases. We follow the approach of \citet{tapas} and generate the reference answer from the reference SQL provided using our own SQL implementation.

\noindent \textbf{\scoder{}} \cite{chen2021spreadsheetcoder} is a dataset to evaluate formula prediction task containing over $17$K spreadsheets extracted from the Enron email corpus that contains $218.798$ instances. It focuses on formula with referenced cells in a rectangular neighbourhood region of the target cell and the headers. We preprocess the data as described Appendix~\ref{sec:appendixC}.

 
\noindent \textbf{\totto} \cite{totto} is a Table-to-Text dataset containing $120.761$ instances. It consists of tables paired with table-grounded sentences as natural language descriptions. \citeauthor{totto} apply several heuristics to sample tables and candidate sentences from Wikipedia pages. They use crowd worker annotators to highlight the corresponding table cells and revise natural language descriptions. 

\noindent \textbf{\finqa{}} \cite{chen2021finqa} is a dataset containing $8.281$ financial Table-QA pairs, along with their numerical reasoning processes in the form a of sequence of mathematical operations. 

\subsection{Results}

We discuss the experimental setup in the Appendix~\ref{sec:appendixB}. For \totto{}, we report the results in Table~\ref{table:tottoresults}. We follow \citeauthor{totto}'s official script to compute BLEU and PARENT as the evaluation metrics. The \emph{Non-Overlap} dev set features examples that are out-of-domain from the training set. For the test set, we provide results from one run as this is a laborious manual process requiring a submission of test files into an external source\footnote{The details on submissions for the ToTTo test set can be found in \href{https://github.com/google-research-datasets/ToTTo}{https://github.com/google-research-datasets/ToTTo}}. Note that \citeauthor{totto} do not provide development set results in their paper and \citeauthor{kale2020text} do not provide test set results for the base model\footnote{The gap between T5-base and \citeauthor{kale2020text} comes from using distinct versions of T5. 
We reproduced their results using v$1.0$. 
%
We use v$1.1$ with the same set of hyperparameters.}. 
We observe that \algo{} outperforms SOTA models and its performance is improved further by using the \textsc{ToTTify} pre-training. 
Note that the base model performs slightly better than the large model. We believe that the large model requires more careful hyperparameters tuning to achieve higher results.
For \wikisql{} and \scoder{}, results are reported in Table~\ref{table:wikisqlresults} and Table~\ref{table:scoderresults-noformula} respectively. Also, see the Appendix~\ref{sec:appendixC} for additional results on the \scoder{} dataset. We observe that \algo\ significantly improves over SOTA performance for both \wikisql{} ($>30\%$ of error reduction in the base variant) and \scoder{} ($35\%$ error reduction in the base variant). Note that \algo\ in the base variant ($220$M parameters) outperforms other models with substantially higher number of parameters (e.g. BERT used in \citet{tapas}
has $380$M parameters and BART$_{large}$ in \citet{liu2022tapex} $ \sim 418$M). Additionally, \algo\ in the small variant ($60$M parameters) achieves high accuracy compared to SOTA for the \scoder{} dataset. When increasing the model size, we observe an increase in performance for both datasets. For \wikisql{} the large variant ($770$M parameters) achieves exceptionally high sequence accuracy of $95\%$  ($53\%$ error reduction wrt. to the baseline performance). 

For \finqa{} we included the top $5$ retrieved passages as part of the input for \algo{}. Additionally we implemented a special tokenization scheme breaking all numbers into single tokens, following~\citet{chowdhery2022palm}. We report the results in table~\ref{table:finqaresults}. The dataset was included as part of the SUKI\footnote{\url{https://suki-workshop.github.io}} workshop~\citep{suki-2022-structured} where \algo{} reached the third place.

\begin{table}[!ht]
\centering
\footnotesize
\begin{tabular}{@{}lS[table-format=2.2(1), separate-uncertainty, detect-weight]S[table-format=2.2(1), , separate-uncertainty, detect-weight]@{}} 
\toprule
\textbf{Model} & \textbf{Dev} & \textbf{Test} \\ [0.5ex] 
\midrule
\citet{tapas}           & 85.1 & 83.6 \\
\citet{liu2022tapex} & 89.2 & 89.5 \\
T5-base  & 85.29(45) & 84.27(39)\\
\hline
\algo-small & 90.56(15) & 89.15(10) \\
\algo-base                   & 92.55(23) & 91.45(21) \\ 
\Hquad \textsc{+ToTTify}  & 91.34(17) & 90.06(15)\\
\Hquad \textsc{-denoising} & 88.87(31) & 87.51(19) \\
\Hquad \textsc{-embeddings} & 85.51(23) & 84.39(13)\\
\algo-large & \B 94.92(4) & \B 93.61(09) \\
\bottomrule
\end{tabular}
\caption{Table-QA results on \wikisql{} in the weakly supervised setting without logical forms. \algo{} provides gains over existing approaches even in a small model variant. The large model gives the best results.} 
\label{table:wikisqlresults}
\end{table}

\begin{table}[!ht]
\centering
\footnotesize
\begin{tabular}{@{}lS[table-format=2.2(1), separate-uncertainty, detect-weight]@{}} 
\toprule

\textbf{Model} & \textbf{Top-1} \\
\midrule
\citeauthor{chen2021spreadsheetcoder} & 42.51\\
\citeauthor{cheng2021fortap}          & 56.30 \\
T5-base & 69.40(33)\\
\hline
\algo-small & 71.33(24)\\
\algo-base & 71.61(27)\\
\Hquad \textsc{+ToTTify} & 71.18(22)\\
\Hquad \textsc{-denoising} & 70.47(28)\\
\Hquad \textsc{-embeddings} & 70.07(36)\\
\algo-large & \B 71.79(20)\\ 
\bottomrule
\end{tabular}
\caption{Formula prediction results on \scoder{}. The T5-base baseline brings substantial improvements over existing approaches. \algo{} provides further gains, with the large model variant obtaining the best results.} 
\label{table:scoderresults-noformula}
\end{table}

\begin{table}[!ht]
\centering
\footnotesize
\begin{tabular}{@{}lS[table-format=2.2(1), separate-uncertainty, detect-weight]S[table-format=2.2(1), , separate-uncertainty, detect-weight]@{}} 
\toprule

\textbf{Model} & \textbf{Program Acc.} & \textbf{Execution Acc.} \\
\midrule
\citeauthor{chen2021finqa} & 61.24 & 58.86\\
T5-base & 62.69 & 60.33 \\
\hline
\algo-base & 66.7 & 64.43 \\
\algo-large & \B  70.79 & 68.00 \\
\bottomrule
\end{tabular}
\caption{Results on the \finqa{} challenge, with an ensemble over $5$ model execution outputs. \algo{} brings substantial improvements over the baselines.}
\label{table:finqaresults}
\end{table}

\begin{figure*}[ht]
    \centering
    \includegraphics[width=1.0\linewidth]{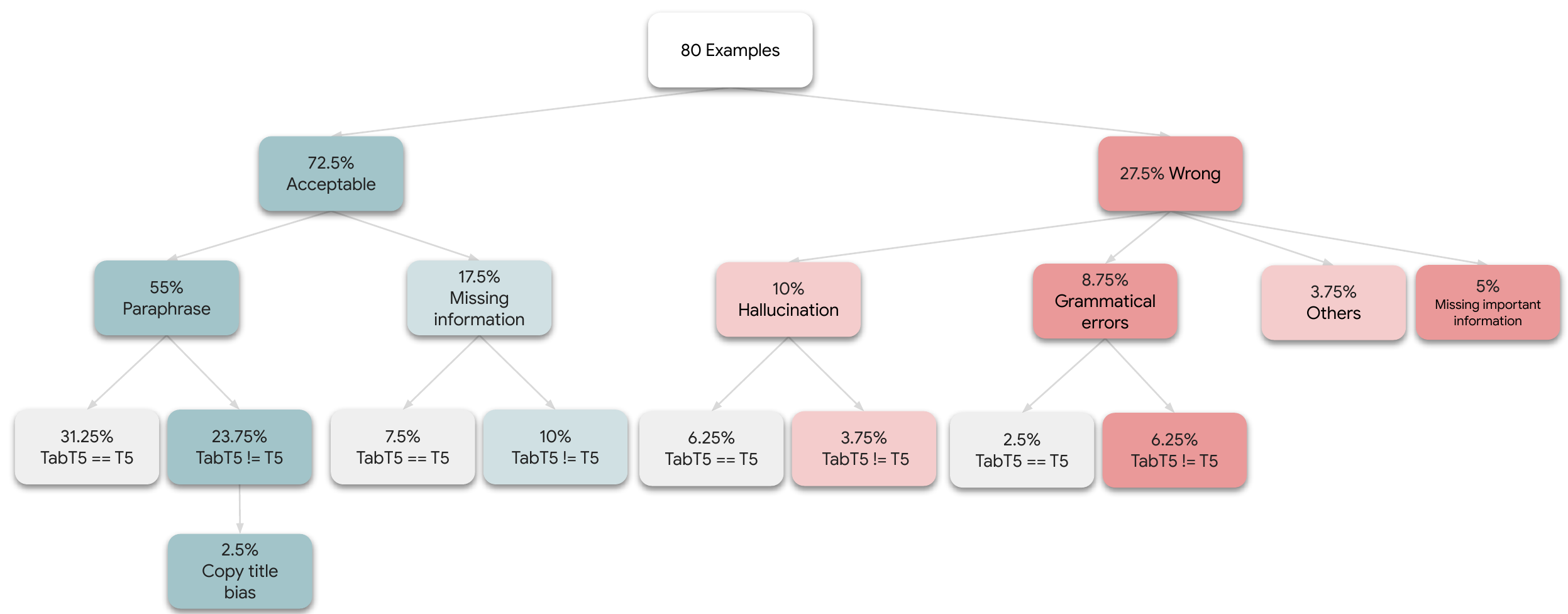}
    \caption{We manually annotate 80  errors made by \algo{}. We find that $55\%$ of predictions are paraphrases 
and $72.5\%$ are acceptable. The classification of error types is given in Table~\ref{table:error_types}.}
    \label{fig:errors_types}.
\end{figure*}


\noindent \textbf{Ablation} We perform the ablation study only on the base variant due to computational costs. For each experiment, we report two ablations runs: (i) \textsc{-denoising} indicates that we remove the denoising pre-training 
, and (ii) \textsc{-embeddings} indicates runs without row and column embeddings.
We observe that the performance of \algo\ deteriorates when removing either denoising or embeddings. This shows they are crucial for all tasks. We also show results for the \textsc{ToTTi}fication, which improves the performance for \totto. 
This pre-training method was defined to imitate the \totto\ task. Thus, it is not surprising that it improves the performance on that task. However, we observe the decrease in performance for other tasks when using this method of pre-training compared to the denoising method. 

\noindent \textbf{Error analysis} We manually annotate a random sample of $80$ errors made by \algo\ on \totto{} dataset, and classify them in Figure~\ref{fig:errors_types}.  

We find that $35\%$ of the \algo{} output are exactly the same as T5’s output where $55\%$ are correct (paraphrases) 
and $72.5\%$ overall are acceptable (correct content with some details missing). 
We classify the remaining errors into grammatical errors, hallucinations and wrong answers (see Appendix~\ref{sec:appendixD} for the errors' definitions and examples). 



The grammatical error cases are mostly around wrongly inserting  determiners like in front of named entities (e.g. \emph{residing in *the* Watauga County}).
Similarly about 50\% of hallucination cases are not \emph{severe}, as they are fluent and convey a similar meaning to the ground truth, but are factually incorrect because the wrong entity being used (e.g. \emph{Washington Wizards} instead of \emph{Washington Huskies}, even when such entities are not in the input). The other 50\% cases, are far from the ground truth meaning.
We speculate that most of the less severe hallucinations cases are directly
connected to the table denoising pretraining scheme employed. That is the
model is biased towards generating related named entities even when those are
not present in the input (i.e. masked during pretraining).

In general, we see that \algo{} is able to produce more fluent sentences than the baselines, either removing superfluous information or using the correct verbs in context (e.g. \emph{served as speaker} instead of \emph{was a speaker}).

The results suggest a need for better metrics for the data-to-text generation tasks that capture the similarities.


\section{Conclusions and Discussion}

We introduced \algo\, a new T5-based encoder-decoder model that improves the encoding ability of tables for pre-trained sequence-to-sequence language model. \algo\
achieves new SOTA results on spreadsheet formula prediction, question answering with text or complex mathematical formulas, and data-to-text generation. 
 This work opens up different paths for future work.
We plan to explore different datasets for pre-training. \citeauthor{t5} show that it is beneficial for the unstructured data to train on datasets bigger than Wikipedia. 
Thus, we plan to use larger and task specific datasets for pre-training (e.g. scrape tables from Web, sheets).
Finally, we want to extend this work to multiple languages, especially low resource ones. 

\newpage
\section*{Limitations \& Ethical Considerations}

As is true for existing works on generative architectures based on large language models, there are potential risks and harms associated with using the output for downstream applications~\cite{bender-koller-2020-climbing, gpt3}. Beyond the original pre-trained checkpoint from T5, we also used tables from Wikipedia for intermediate pre-training, which may contain additional undesirable biases.

\section*{Acknowledgements}

We would like to thank Massimo
Nicosia, Dima Brezhnev and Ankur Parikh, as well as the anonymous reviewers for their constructive
feedback, useful comments and suggestions.

\bibliography{bib/journal,bib/custom}
\bibliographystyle{acl_natbib}

\clearpage
\newpage
\appendix

\section{Hyperparameters Selection}
\label{sec:appendixA}

We run denoising pre-training for $1$M steps and ToTTify pre-training for $100$k steps on top of denoising. We set each fine-tuning task for $50$k training steps. We run the evaluation on \scoder{} and \wikisql{} using the default T5 hyper-parameters with an input sequence length of $1024$ and output $256$. For the \totto{} dataset, we follow the approach of \citet{kale2020text} and keep the learning rate constant and equal to \SI{1e-4}, an input and output sequence length is equal to $512$ and $256$ respectively, and batch size is $256$. Additionally, we observe that \algo\ in the small and base variants overfit quickly. Thus, we decide to increase the dropout rate to 0.2 when using pre-training. 

\section{Experimental Setup}
\label{sec:appendixB}

We apply the standard T5 tokenizer and start pretraining from publicly available T5 checkpoints. Row and column embeddings are randomly initialized.
We run pre-training and fine-tuning on a setup of $16$ Cloud TPU v3 cores with maximum sequence length of $1024$. Pre-training takes around $3$, $8$ and $13$ days for small, base and large models. Fine-tuning takes around $2-3$ hours for each task. For each dataset, we run five independent runs and report median and standard deviation.

\section{\scoder{} results.}
\label{sec:appendixC}

In this section, we present the results on the \scoder{} dataset that contains all original data (i.e. all formulas in the tables). We find these results interesting as the \scoder{} contains real data collected by the company. Thus, we believe this scenario is realistic. We present the results in Table~\ref{table:scoderresults}. 
\begin{table}[ht]
\centering
\begin{tabular}{@{}lS[table-format=2.2(1), separate-uncertainty, detect-weight]@{}} 
\toprule

\textbf{Model} & \textbf{Top-1} \\
\midrule
T5-base & 93.05(98)\\
\hline
\algo-small & 95.39(17)\\
\algo-base & \B 95.59(8)\\
\Hquad \textsc{+ToTTify} & 95.50(5)\\
\Hquad \textsc{-denoising} & 93.92(16)\\
\Hquad \textsc{-embeddings} & 95.00(16)\\
\bottomrule
\end{tabular}
\caption{Formula prediction results on \scoder{}. In this experiment, the model has to produce the target formula having access to the formula used in the surrounding cells. Results are higher wrt. to Table~\ref{table:scoderresults-noformula} as the model is allowed to ``copy'' already used formulae or part of them.} 
\label{table:scoderresults}
\end{table}
We observe that our results are extremely high because in \scoder{} dataset over $70\%$ of tables contain a target formula in the input table. Following the previous approaches, we make the task harder. In the experimental section of the paper, we preprocess the datasety by removing all formulas from the input table cells. Additionally, we remove examples containing (i) erroneous formulas, and (ii) ranges from different tables in both input tables and target formulas.

\begin{table*}[ht]
\resizebox{\textwidth}{!}{
\begin{tabular}{p{5cm} p{8cm}}

\textbf{Error type} & \textbf{Definition} \\
\toprule
Paraphrase & Express the same meaning as the ground truth using either synonyms or the exact words in a different order. \\
\multicolumn{2}{p{13cm}}{
\vspace{-0.25cm}
\small
\begin{itemize}[noitemsep]
    \item \algo{}: \textit{Ina 2016, Alma Jodorowsky played Evelyn in Kids in Love.}
    \item \gold{}: \textit{Alma Jodorowsky had the role of Evelyn in 2016 film Kids in Love.}
\end{itemize}
}
\\

\toprule
Acceptable \newline (\emph{missing information}) & The content is correct, but it is missing some details that do not affect the answer's meaning. \\
\multicolumn{2}{p{13cm}}{
\vspace{-0.25cm}
\small
\begin{itemize}[noitemsep]
    \item \algo{}: \textit{The 500 Questions was aired in Germany on RTL from July 4 to August 14, hosted by Günther Jauch.} (year is missing)
    \item \gold{}: \textit{In 2016, RTL television aired 500-DQA in germany and was hosted by Gunther Jauch.}
\end{itemize}
}
\\

\toprule
Wrong \newline (\emph{missing important information}) & The content is missing some details that affect the meaning of the answer or are essential for understanding the answer.\\
\multicolumn{2}{p{13cm}}{
\vspace{-0.25cm}
\small
\begin{itemize}[noitemsep]
    \item \algo{}: \textit{Putney railway station is in the Wandsworth borough and is in Zone 2.} (missing zone 3 could be important information)
    \item \gold{}: \textit{Putney railway station serves Putney in the London borough of Wandsworth in southwest London and in zones 2 and 3}.
\end{itemize}
}
\\

\toprule
Hallucination &
\vspace{-0.25cm}
\begin{itemize}[noitemsep,topsep=0pt,leftmargin=*]
    \item Intrinsic: output contradicts the source content.
    \item Extrinsic: output cannot be verified from the source content.
\end{itemize}\\
\multicolumn{2}{p{13cm}}{
\vspace{-0.25cm}
\small
\begin{itemize}[noitemsep,topsep=0pt]
    \item \algo{}: \textit{In 1924, William Glackens received the Temple Gold Medal for his work "Natural form".} (We cannot verify the work's name)
    \item \gold{}: \textit{William Glackens won the 1924 award from Temple Gold Medal Nude.}
\end{itemize}
}
\\

\toprule
Grammatical errors & The sentence is ungrammatical.\\
\multicolumn{2}{p{13cm}}{
\vspace{-0.25cm}
\small
\begin{itemize}[noitemsep]
    \item \algo{}: \textit{As of the census of 2000, there were 42,695 people residing in the Watauga County.}
    \item \gold{}: \textit{As of the census of 2000, there were 42,695 people residing in Watauga county.}
\end{itemize}
}
\\
\toprule
Wrong & Other errors such as: wrong aggregation (counts, sums, etc), swapped arguments that change the meaning of the sentence.\\
\toprule
\algo{} == T5 & T5 and \algo{} have the same output verbatim (exact match). \\
\bottomrule
\end{tabular}
}
\caption{Error types definition and examples.}
\label{table:error_types}
\end{table*}

\section{Error analysis}
\label{sec:appendixD}

We manually annotate  $80$ errors made by the \algo{} on \totto{} dataset, and classify them according to the definitions in Table~\ref{table:error_types}. 

\end{document}